\documentclass[sigconf]{acmart}

\makeatletter
\renewcommand\@formatdoi[1]{\ignorespaces}
\makeatother
\usepackage{amsthm}
\usepackage{graphicx, subfigure}
\usepackage{caption}
\usepackage{epstopdf}
\usepackage{mathtools}
\usepackage{hyperref}
\usepackage{nameref}
\usepackage{enumerate}
\usepackage{paralist}
\usepackage{tabularx}
\usepackage{float}

\usepackage[ruled,vlined,algo2e]{algorithm2e} 
\newcommand{\figwidth}{.9}
\newcommand{\subfigwidth}{.49}
\newcommand{\threesubfigwidth}{.325}

\newcommand{\boxscalenodeorder}{.80}
\newcommand{\bannerwidth}{1}
\newcommand{\boxscale}{.9}
\newcommand{\boxscalemodelselect}{.8}
 
\DeclareMathOperator*{\iqr}{iqr}
\DeclareMathOperator*{\median}{med}

\DeclareMathOperator*{\cost}{bytes}
\DeclareMathOperator*{\correct}{correct}

\DeclareMathOperator*{\argmax}{argmax}
\SetKwInput{KwGiven}{Given}
\SetKwInput{KwFind}{Select}
\SetKwInput{KwWhere}{Where}
\SetKwInput{KwMin}{Minimizing}
\newcommand{\bfs}{\texttt{bfs}}
\newcommand{\cluster}{\texttt{cluster}}
\newcommand{\degree}{\texttt{degree-net}}
\newcommand{\activity}{\texttt{activity-net}}
\newcommand{\random}{\texttt{random}}
\newcommand{\knns}{$\texttt{KNN}_s$}

\newcommand{\networkquery}{\mathcal{Q}}
\newcommand{\networkqueryr}{\mathcal{Q}_r}
\newcommand{\classifier}{\mathcal{C}}
\newcommand{\classifierr}{\mathcal{C}_r}
\newcommand{\classifierset}{C_r[i,k]}
\newcommand{\degreeflat}{\texttt{degree-top}}
\newcommand{\activityflat}{\texttt{activity-top}}
\newcommand{\ba}{BeerAdvocate}
\newcommand{\lfm}{Last.fm}
\newcommand{\ml}{MovieLens}
\newcommand{\knn}{\texttt{KNN}}
\newcommand{\tth}{\texttt{TH}}
\newcommand{\social}{\texttt{social}}
\newcommand{\bigo}{\mathcal{O}}
\newcommand{\netcost}{\cost(\networkqueryr)}
\newcommand{\taskcost}{\cost(\classifierr)}
\newcommand{\taskcorrect}{\correct(\classifierr)}
\newcommand{\efficiency}{\mathcal{E}(\networkqueryr)}

\SetAlFnt{\small}
\SetAlCapFnt{\small}
\SetAlCapNameFnt{\small}

\begin{document}

\title{Network Model Selection Using Task-Focused Minimum Description Length}

\author{Ivan Brugere}
\affiliation{%
  \institution{University of Illinois at Chicago}
  \city{Chicago}
  \state{IL}}
\email{ibruge2@uic.edu}

\author{Tanya Y. Berger-Wolf}
\affiliation{%
  \institution{University of Illinois at Chicago}
  \city{Chicago}
  \state{IL}}
\email{tanyabw@uic.edu}

\begin{abstract} Networks are fundamental models for data used in practically every application domain. In most instances, several implicit or explicit choices about the network definition impact the translation of underlying data to a network representation, and the subsequent question(s) about the underlying system being represented. Users of downstream network data may not even be aware of these choices or their impacts. We propose a task-focused network model selection methodology which addresses several key challenges. Our approach constructs network models from underlying data and uses minimum description length (MDL) criteria for selection. Our methodology measures \textit{efficiency}, a \textit{general} and comparable measure of the network's performance of a local (i.e. node-level) predictive task of interest. Selection on efficiency favors parsimonious (e.g. sparse) models to avoid overfitting and can be applied across arbitrary tasks and representations. We show \textit{stability, sensitivity, and significance testing} in our methodology.\end{abstract}
\acmDOI{}
\settopmatter{printacmref=false} % Removes citation information below abstract
\renewcommand\footnotetextcopyrightpermission[1]{} % removes footnote with conference information in first column
\pagestyle{plain} % removes running headers
\maketitle

% Copyright Statement
% When submitting your final paper to a SIAM proceedings, it is requested that you include 
% the appropriate copyright in the footer of the paper.  The copyright added should be 
% consistent with the copyright selected on the copyright form submitted with the paper.
% Please note that "20XX" should be changed to the year of the meeting.

% Default Copyright Statement
% \fancyfoot[R]{\footnotesize{\textbf{Copyright \textcopyright\ 20XX by SIAM\\
% Unauthorized reproduction of this article is prohibited}}}

% Depending on which copyright you agree to when you sign the copyright form, the copyright 
% can be changed to one of the following after commenting out the default copyright statement
% above.

%\fancyfoot[R]{\footnotesize{\textbf{Copyright \textcopyright\ 20XX\\
%Copyright for this paper is retained by authors}}}

%\fancyfoot[R]{\footnotesize{\textbf{Copyright \textcopyright\ 20XX\\
%Copyright retained by principal author's organization}}}

%\pagenumbering{arabic}
%\setcounter{page}{1}%Leave this line commented out.

%Do these choices yield a model which is actually predictive of the underlying system? Can we discover better (i.e. more predictive, simpler) models?

\section{Introduction}

Networks are fundamental models for data used in practically every application domain. In most instances, several implicit or explicit choices about the network definition impact the translation of underlying data to a network representation, and the subsequent question(s) about the underlying system being represented. Users of downstream network data may not even be aware of these choices or their impacts. Do these choices yield a model which is actually predictive of the underlying system? Can we discover better (i.e. more predictive, simpler) models? %Our work  For example, simple rule-based or threshold-based criteria define an edge between entities. 

A network derived from some data is often assumed to be a true representation of the underlying system. This assumption introduces several challenges. First, these relationships may be very time-dependent: an aggregation of edges might not reflect the behavior of the underlying system at \textit{any} time \cite{Caceres2013}. Second, relationships may be context-dependent: where the same network inferred from data predicts one behavior of the system but not another \cite{brugereicdm2017}. Any subsequent measurement on the network (e.g. degree distribution, diameter, clustering coefficient, graph kernels) treats all edges equally, when a path or community in the network may only be an error of aggregation.

Task-based network model selection addresses the latter challenge by comparing multiple representations for predicting a particular behavior (or context) of interest. In this context, the network is a space which structures a local predictive task (e.g. ``my neighbors are most predictive of my behavior''), and model selection reports the best structure. 

Work in network model selection has typically focused on inferring parameters of generative models from a given network, according to some structural or spectral features (e.g. degree distribution, eigenvalues) \cite{10.1371/journal.pone.0049949}. However, preliminary work has extended model selection to criteria on task performance with respect to generative models for a given network \cite{2016arXiv160904859C}, and methodologies representing underlying data as networks \cite{brugereicdm2017}.

We propose a task-focused network model selection methodology which addresses several key challenges. Our approach constructs network models from underlying data and uses minimum description length (MDL) criteria for selection. Our methodology measures \textit{efficiency}, a \textit{general} and comparable measure of the network's performance of a local (i.e. node-level) predictive task of interest. Selection on efficiency favors parsimonious (e.g. sparse) models to avoid overfitting and can be applied across arbitrary tasks and representations. We show \textit{stability, sensitivity, and significance testing} in our methodology.

\begin{figure*}[t]
\centering
  \includegraphics[width=\bannerwidth\textwidth]{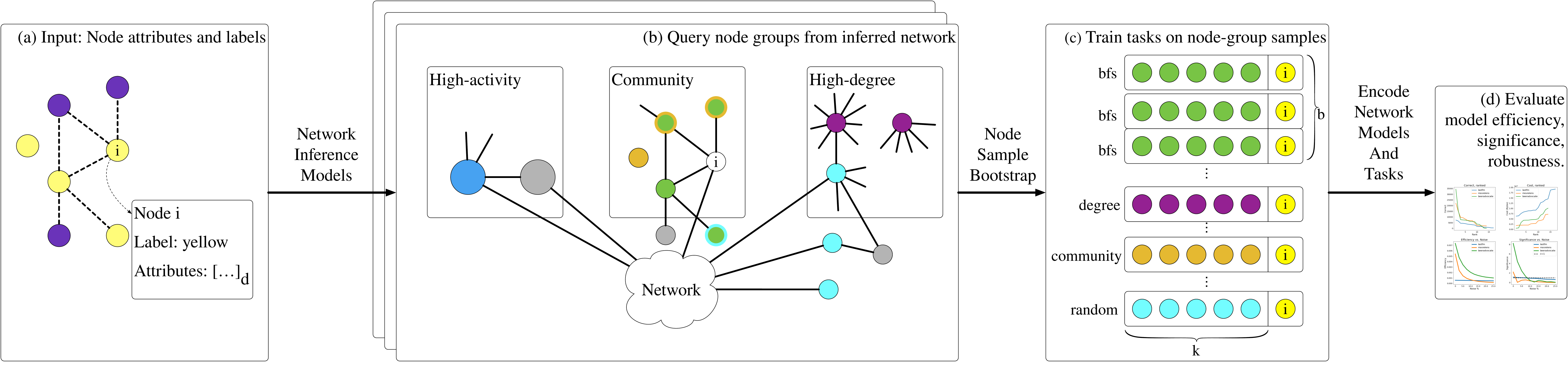}
  \caption{A high-level overview of our methodology. {\bf (a)} Given as input: nodes with associated attribute and label data, potential network structure(s) are optionally provided (dashed lines). We use network inference models to construct multiple networks from data (note: stack of models). {\bf (b)} On each network model, for each node `$i$', we sample nodes (shown by node color) and their associated attributes and labels, according to different sample methods and node characterization heuristics (e.g. `high degree', `high activity'). {\bf (c)} we generate `$b$' samples by each method at varying sample size $k$, each row is a trained task model (e.g. random forest) to predict label `$i$' (`yellow') given sampled attribute data and labels. {\bf (d)} we encode each collection of task models from (c) and network representation from (b) to measure the most \textit{efficient} model for performing the task, and \textit{select} that network representation.} 
  \label{fig:overview}
\end{figure*}

\subsection{Related work}
%in neuroscience \cite{Sporns2014},  bioinformatics \cite{ZhangHorvath2005}, or behavioral ecology \cite{Farine2016}, or online trajectories and user activity in recommender systems \cite{McAuley:2015:INS:2783258.2783381}
Our work is related to network structure inference \cite{2016arXiv161000782B, Kolaczyk2009}. This area of work constructs a network \textit{model} representation for attributes collected from sensors, online user trajectories or other underlying data. Previous work has focused on these network representations as models of local predictive tasks, performed on groups of entities, such as label inference \cite{namata:tkdd15} and link prediction \cite{Hasan2011, Liben-Nowell2007}. A `good' network in this setting is one which performs the task well, under some measure of robustness, cost, or stability. Previous work examined model selection under varying network models, single or multiple tasks, and varying task methods \cite{mlg2017_13, brugereicdm2017}. 

The minimum description length (MDL) principle \cite{10.1002/0471667196.ess1641.pub2} measures the representation of an object or a model by its smallest possible encoding size in bytes. This principle is used for model selection for predictive models, including regression \cite{doi:10.1198/016214501753168398} and classification \cite{Mehta1996}. MDL methods have also been applied for model selection of \textit{data representations} including clusterings, time series \cite{Hu:2011:DIC:2117684.2118274, doi:10.1198/016214501753168398}, networks \cite{doi:10.1093/bioinformatics/btl364} and network summarization \cite{doi:10.1137/1.9781611973440.11}. MDL has been used for structural outlier and change detection in dynamic networks \cite{Ferlez4497545, Sun:2007:GPM:1281192.1281266}. Our methodology encodes a collection of predictive models, together with the underlying network representation for model selection. No known work encodes both of these objects for model selection. 
%Eberle4476697, 

Several generative models exist to model correlations between attributes, labels, and network structure, including the the Attributed Graph Model (AGM) \cite{Pfeiffer:2014:AGM:2566486.2567993}, and Multiplicative Attribute Graph model (MAG) \cite{Kim2012}. These models are additive to our methodology, and could be applied and evaluated as potential models within our methodology.

Our work is orthogonal to graph kernels \cite{Vishwanathan:2010:GK:1756006.1859891} and graph embeddings \cite{2017arXiv170905584H}. These methods traverse network structure to extract path features and represent nodes in a low-dimensional space, or to compare nodes/graphs by shared structural features. We treat this graph traversal as one possible query of nodes for input to local tasks, but do not compare graphs or nodes directly; our focus is to evaluate how well the fixed model performs the local task over all nodes. 
% \todo{add dechoudry MSR (task oriented), manuel cascades (network models)}

%, and work in graph kernels and embeddings. 

%or model selection among many possible representations. Previous work focuses on model selection on  This principle selects 

%Network structure inference

% \cite{}
%   , neuroscience , and recommender systems . Much of this work has domain-driven network model evaluation and lacks a general methodology for transforming data to useful network representations.
% \cite{Gomez-Rodriguez:2012:IND:2086737.2086741, Myers2010}. 

% label prediction:

%   \cite{McDowell:2013:LAR:2505515.2505628} \cite{4476695} \cite{namata:tkdd15}. 

% MDL:
% \cite{}
% \cite{}
% \cite{}

% (Distinction) Graph kernels/embeddings:

% Distinction network modeling:

% Similarity-based methods tend to be ad-hoc, incorporating domain knowledge to set network model parameters. Recent work on task-focused network inference evaluates inferred network models according to their ability to perform a set of tasks \cite{DeChoudhury:2010:IRS:1772690.1772722}. These methods often have a high sensitivity to threshold/parameter choice, and added complexity of interactions between network representations and task models. Our work identifies these sensitivities, and yields robust model selection over several stability criteria.

\subsection{Contributions}

We formulate a task-focused model selection methodology for networks. 

\begin{compactitem}
    \item \textit{Model selection methodology}: We propose a generalized approach for evaluating networks over arbitrary data representations for performing particular tasks. 
    \item \textit{Network efficiency}: We propose a general minimum description length (MDL) \textit{efficiency} measure on the encoding of network representations and task models, which is comparable over varying models and datasets,
    \item \textit{Validation and significance} We empirically demonstrate stability, sensitivity, and significance testing of our model selection methodology.
\end{compactitem}

% model selection framework
%    model measure
% stability, significance testing

%approach
%    
\section{Methods}

Figure \ref{fig:overview} gives a schematic overview of our model selection methodology. In Figure \ref{fig:overview}(a) as input, we are given attribute-set $A$ and label-set $L$ associated with entities $i \in V$. Labels are a notational convenience to denote an attribute of interest for some predictive task (e.g. label inference, link prediction). We \textit{may} be given a network topology (dashed lines), which we treat as one possible network model in the evaluation.

We apply a collection of network inference methods to generate separate network topologies. In Figure \ref{fig:overview}(b), for each network topology, we use several network sampling methods to sample attribute and label data for input to a supervised classification/regression task. Boxes indicate different node characterization heuristics which can be sampled: `high degree', `high activity', or `communities'. Node colors indicate one sample from each of sampling method with respect to node $i$: green nodes denote a \textit{local}, orange nodes denote a \textit{community} sample (the same node in two samples indicated by two-color nodes). Purple is a sample of \textit{high-degree} nodes. Blue represents a sample of high-activity nodes (e.g users with the most interactions, most purchases, most ratings), and cyan represents a random sample of nodes.

In Figure \ref{fig:overview}(c), each network sample method produces samples of length $k$, and is repeatedly sampled to create $b$ supervised classification instances per method. Each row corresponds to a supervised classification task trained on attributes and labels in the sample, to predict the label of $i$ (e.g. `yellow'). Finally, in Figure \ref{fig:overview}(d), we encode the predictive models (e.g. the coefficients of an SVM hyper-plane or random forest trees) and the network in an efficient byte-string representation, and evaluate the best model using minimum description length (MDL) criteria. We propose a general and interpretable measure of \textit{efficiency} for solving our task of interest, and show stability and significance testing over our collection of models. 

%For each $i\in V$, $\vec{a}_i \in A$ are its attributes and $l_i\in L$ is a label to be learned.

%(In a network context, this is equivalent to learning a local model for link prediction or label prediction for $i$.) Our methodology evaluates the efficiency of $\networkquery$ for performing supervised learning task(s) using $\classifier$
\subsection{Preliminaries and notation}
Let $V=\{1...,i...,n\}$ be a set of entities to which we refer to as `nodes.' Let $A=\{\vec{a}_i\}$ be a set of attribute-vectors and $L=\{l_i\}$ be a set of target labels of interest to be learned, where $i \in V$.
Let $\networkquery:V\times \mathbf{Z}^+ \times (\mbox{other data})\to V^k$ be a network query function (see the full definition below) that produces query-sets of nodes $S \subseteq V$ of a given size $k$. Let $\classifier:A\times L\times V \to L$ be a classifier trained on attributes and labels of a node subset $S$ to produce a prediction $l'_i$, for node $i$.\footnote{Throughout, capital letters denote sets, lowercase letters denote instances and indices. Script characters (e.g. $\classifier$) and keywords (e.g. $\cost()$) denote functions, teletype (e.g. \knn-\bfs) denotes \textit{models}, square brackets (e.g. $A[S]$) denote the restriction of a set to a subset.}  %

\subsection{Network query functions}

In our methodology, we access the network for our given objective using a \textit{network query function} on node $i$, which returns a set of nodes of arbitrary size subject to the network topology and function definition. 

To compare different query functions under similar output, we introduce a bounded query of size $k$. With an input of node $i$ and search size $k$, bounded network query function $\networkquery$ outputs a set of  unique nodes:\small{}
\begin{equation}
\networkquery(i, k) \rightarrow \{j_{1}, j_{2},...,j_{k}\}, \text{ where } i \notin\{j_{1},...,j_{k}\}
\label{eq:nodeorder}
\end{equation}
\normalsize{}In general, these queries sample from the possible node-sets, yielding different result-set samples on a repeated input $i$. We measure these functions over a distribution of node-set samples, with replacement, analogous to the bootstrap sample \cite{efron93bootstrap}.

Network \textit{adjacency} is one such query method, which randomly samples from ${\mathcal{N}(i) \choose k}$ subsets, with replacement. Breadth-first search $\networkquery_{\bfs}(i,k,E)$ generalizes adjacency to higher-order neighbors, where nodes in the farthest level is selected at random. Over varying query size $k$, breadth-first search is well-suited for measuring the performance of the network in finding relevant nodes `nearer' to the query. 

While $\networkquery$ may sample an underlying network edge-set $E$, our methodology evaluates any network query function, regardless of underlying implementation. In geometric space, $\networkquery_{l_2}$ may sample nodes in increasing Euclidean distance from $i$. Query heuristics can also be derived from the attribute-set $A$. For example, $\networkquery_{\activity}(i, k, A)$ (defined below) may query points by the number of non-zero attributes or some other attribute function, and sample according to some similarity to $i$.

\subsubsection{Network efficiency}

We evaluate each network query function according to a minimum description length (MDL) criterion. We repeatedly sample $\networkqueryr(i, k)$ ($r$ for `rule'), to construct $b$ samples of attribute vectors and labels, of size $k$. We train each classifier $\classifier(A[S],L[S])$ from these attribute and label data. Let $\classifierset$ be the resulting set of trained classifiers: \small{}
\begin{equation}
\begin{aligned}
\classifierset = \{\classifier(A[\networkqueryr(i,k)_1], L[\networkqueryr(i,k)_1]),...\\
\classifier(A[\networkqueryr(i,k)_b], L[\networkqueryr(i,k)_b])\}_{|b|}
\end{aligned}
\end{equation}
\normalsize{}Let $\correct(\classifierset)$ be the sum of correct predictions of the $b$ task classifiers on test input $\classifierr(\vec{a_i}, l_i)$. 
\begin{definition}
The {\em efficiency} of a network query function $\networkqueryr$ with respect to a node $i$ is the number of correct predictions per byte (higher is better). The efficiency $\mathcal{E}$ is the maximum given by the value of $k$ which yields the maximum of correct predictions divided by the median encoding cost, both aggregates of $b$ samples:\small{}
\begin{equation}
\mathcal{E}(\networkqueryr, i) = \max_k\left\{{\frac{\correct(\classifierset)}{\median_b(\cost(\classifierset))}}\right\}. 
\label{eq:efficiency}
\end{equation}
\end{definition}
\normalsize{}Let $\kappa_i=\argmax_{k}\mathcal{E}(\networkqueryr, i)$, the $k$ associated with the maximum efficiency of node $i$. We then sum $\cost(\classifierr[i,\kappa_i])$ and $\correct(\classifierr[i,\kappa_i])$ over all $i$ to get efficiency over task models trained on samples from $\networkqueryr$. We also encode $\networkqueryr$ and measure its \textit{representation} cost:

%Over all $i$, there is a distribution of $\kappa_i$ values \todo{show}.   

% $\cost(\mathcal{C}_r[i,*])$ and $\correct(\mathcal{C}_r[i,*])$ denote the quotient factors in Equation \ref{eq:efficiency} (median bytes, number correct) associated with the $k$ with minimum inefficiency for node $i$. . 

\small{}
\begin{equation}
\begin{split}
\taskcorrect{} = \sum_i\correct(\classifierr [i, \kappa_i])\\
\taskcost{} =\sum_i\cost(\classifierr [i,\kappa_i])
\end{split}
\label{eq:efficiencyseparate}
\end{equation}
\normalsize{}Then, the overall efficiency of $\networkqueryr$ is given by:\small{}
\begin{equation}
\efficiency = \frac{\taskcorrect}{\taskcost + \netcost}. 
\label{eq:efficiencyfull}
\end{equation}
\normalsize{}Encoding the function $\networkqueryr$ favors parsimonious models and avoids overfitting the representation to evaluation criteria such as accuracy or precision.  However, such encoding requires careful consideration of the underlying representation, which we cover in Section \ref{subsubsec:nodeordering}. Omitting the encoding cost of $\networkqueryr$ in Equation \ref{eq:efficiencyfull} measures only the extent to which the network query function trains `good' task models in its top-$k$ samples. This is suitable if we are not concerned with the structure of the underlying representation, such as its sparsity. 
To favor interpretability of efficiency in real units (correct/byte), we avoid weight factors for the terms in the efficiency definition. However, in practice this would provide more flexibility to constrain one of these criteria. 

\subsection{Problem statement}

We can now formally define our model selection problem, including inputs and outputs:

\begin{algorithm2e}%[H]
  \SetAlgorithmName{Problem}
  \\
  \\
  \KwGiven{A set of nodes $V$, Node Attribute-set $A$, Node Label-set $L$,
  Network query function-set  $Q=\{\networkqueryr\}$, where $\networkqueryr (i,k) \rightarrow \{v_{j_1}, v_{j_2}, ...v_{j_k}\}$, Task $\classifier$, where $ \classifier(A[S],L[S], \vec{a}_i) \rightarrow l'_i$}
  \KwFind{Network query function $\networkquery' \in Q$}
  \KwWhere{$\networkquery' = \argmax_r \efficiency$}
  \caption{MDL Task-Focused Network Inference Model Selection}
  \label{p:networkinference}
\end{algorithm2e}

For brevity, we refer to `model selection' simply as selecting the network representation and its associated network query function for our task of interest. The underlying query functions in set $Q$ may be implemented as arbitrary representations which can be randomly sampled (e.g. lists). This model selection methodology measures, for example whether a network is a better model than other non-network sampling heuristics, group-level sampling (e.g. communities), etc. accounting for representation cost of each. That is, whether a network is necessary in the first place. As a consequence, we may select a different queried representation than a network.

Previous work has evaluated network model-selection for robustness to multiple tasks (e.g. link prediction, label prediction) as well as different underlying task models (e.g. random forests, support vector machines) \cite{brugereicdm2017}. Problem \ref{p:networkinference} can straightforwardly select over varying underlying network models, tasks, or task models which maximize efficiency. We simplify the selection criteria to focus on measuring the efficiency of network query functions and their underlying representations, but this current work is complimentary to evaluating over larger parameter-spaces and model-spaces. 

\subsection{Network models}

We define several networks queried by network query functions $\networkqueryr$. Our focus is not to propose a novel network inference model from attribute and label data (see: \cite{Pfeiffer:2014:AGM:2566486.2567993,Kim2012}). Instead we apply existing common and interpretable network models to demonstrate our framework. The \textit{efficiency} of any novel network inference model should be evaluated against these baselines. 

A network model $\mathcal{M}$ constructs an edge-set from node attributes and/or labels: $\mathcal{M}_j: \mathcal{M}_j(A,L) \rightarrow E_j$. We use simple $k$-nearest neighbor (\knn) and Threshold (\tth) models common in many domains. 

Given a similarity measure $\mathrm{d}(\vec{a}_i, \vec{a}_j) \rightarrow s_{ij}$ and a target edge count $\rho$, this measure is used to produces a pairwise attribute similarity space. We select edges by:

%\footnote{Sub-quadratic approximation of these similarity spaces vary by measure, and is outside the scope of our methodology.}

\begin{compactitem}
	\item $k$-nearest neighbor $\mathcal{M}_{\knn}(A, \mathrm{d}(), \rho)$: for a fixed $i$, select the top $\lfloor{\frac{\rho}{|V|}}\rfloor$ most similar $\mathrm{d}(\vec{a}_i, \{A \setminus \vec{a}_i\})$. In directed networks, this produces a network which is $k$-regular in out-degree, with $k=\lfloor{\frac{\rho}{|V|}}\rfloor$.
	\item Threshold $\mathcal{M}_{\tth}(A, \mathrm{d}(), \rho)$: over all pairs $(i,j)\in V$ select the top $\rho$ most similar $\mathrm{d}(\vec{a}_i, \vec{a}_j)$.
\end{compactitem} 

Let these edge-sets be denoted $E_{\knn}$ and $E_{\tth}$, respectively. We use varying network sparsity ($\rho$) on these network models to choose `sparse' or `dense' models.
Similarity measures may vary greatly by application. In our context, attribute data are counts and numeric ratings of items (e.g. films, music artists) per user. We use a sum of intersections (unnormalized), which favors higher activity and does not penalize mismatches:\small{}
\begin{equation}
    \begin{aligned}
	\mathrm{d}_{INT}(\vec{a}_i, \vec{a}_j) = \sum\nolimits_l \mathrm{min}(a_{il}, a_{jl})
	\end{aligned}
\end{equation}
\normalsize{}\subsubsection{Encoding networks and tasks}
\label{subsubsec:nodeordering}
Table \ref{tab:functions} summarizes the network query functions (Equation \ref{eq:nodeorder}) used in our methodology, and their inputs. We define only a small number of possible functions, focusing on an interpretable set which helps characterizes the underlying data. The encoding cost increases from top to bottom.\footnote{For simplicity, we refer to models only by their subscript labels, e.g. \knns-\bfs{} for the sparse $\networkquery_{\bfs}(E_{\knn})$ model.}

\activityflat{} and \degreeflat{} are the simplest network query functions we define. \activityflat{} sorts nodes by `activity', i.e. the number of non-zero attributes in $\vec{a}_i$, and selects the top-$\ell$ ranked nodes ($\ell << |V|$). An ordering is generated by random sub-sampling of this list. \degreeflat{} does similar on node degree with respect to some input $E$.

\cluster{} applies graph clustering (i.e. community detection) on the input edge-set $E$. This reduces the network to a collection of lists representing community affiliation. Each node is affiliated with at most one community label. Although both \degreeflat{} and \cluster{} are derived from an underlying network, we only encode the reduced representation. This is by design, to measure whether the network is better represented by a simple ranking function which we represent in $\bigo(\ell)$ space, a collection of groups in $\bigo(|V|)$ space, or edges in $\bigo(|E|)$ space. \random{} produces random node subset, encoded in $\bigo(|V|)$ space. 

\bfs{} is a breadth-first search of an underlying edge-set $E$ from seed $i$. This is encoded by an adjacency list (a list of lists), in $\bigo(|E|)$. \degree{} is an ad-hoc network with all out-edges to some node in the top-$\ell$ nodes of \degreeflat{}. For each node $i$, we select $m < \ell$ most similar nodes and define out-edges from $i$. This is a \knn{} graph, where $k=m$, constrained to high-degree nodes. This additional encoding cost relative to \degreeflat{} measures the extent that specialization exists in the top-ranked individuals as a set of task `exemplars' for all nodes. \activity{} is defined analogously with respect to \activityflat{}.

\small{}
\begin{table}
\centering
\resizebox{\boxscalenodeorder\columnwidth}{!}{
\begin{tabular}{|p{.5\columnwidth}|p{.20\columnwidth}|}
\hline
Network Query Function & Encoding\\
\hline \hline
$\networkquery_{\activityflat}(i,k,\mathrm{sort}(A),\ell)$  & $\bigo(\ell)$\\
\hline
$\networkquery_{\degreeflat}(i,k,\mathrm{sort}(E),\ell)$  & $\bigo(\ell)$\\
\hline
$\networkquery_{\cluster}(i,k,\mathrm{cluster}(E))$ & $\bigo(|V|)$ \\
\hline
$\networkquery_{\random}(i,k,V)$  & $\bigo(|V|)$\\
\hline
\hline
$\networkquery_{\bfs}(i,k,E)$ & $\bigo(|E|)$ \\
\hline
$\networkquery_{\activity}(i,k,\mathrm{sort}(A),m,\ell)$  & $\bigo(|V|\times m)$\\
\hline
$\networkquery_{\degree}(i,k,\mathrm{sort}(E),m,\ell)$  & $\bigo(|V|\times m)$\\
\hline
\end{tabular}}
\caption{The input signature and space complexity of network query functions. We define four functions implemented by non-network structures (top), and three implemented by a network (bottom)}
\label{tab:functions}
\end{table}
\normalsize{}
To demonstrate our methodology, we use Random Forest task models. Each decision tree is encoded as a tuple of lists which represent left and right branches, associated feature ids, and associated decision thresholds. The random forest is then a list of these individual tree representations.

\subsection{Measuring efficiency}

All of our network query functions and task models can now be represented as a byte-encodable object `o' (e.g. lists). Therefore, we can now measure efficiency (Equation \ref{eq:efficiencyfull}). To implement our $\cost(o)$ function estimating the minimum description length, we convert each object representation to a byte-string using the msgpack library,\footnote{\url{https://msgpack.org/}} We then use LZ4 compression,\footnote{\url{https://lz4.github.io/lz4/}} (analogous to zip) which uses Huffman coding to reduce the cost of the most common elements in the representation string (e.g high degree nodes, frequent features in the random forest). Both of these methods are chosen simply by necessity for runtime performance. Finally, we report the length of the compressed byte-string:
\small{}
\begin{equation}
    \cost(o) = |\mathrm{lz4.dumps}(\mathrm{msgpack.dumps}(o))|	
\end{equation}
\normalsize{}\subsubsection{Network query reach}
\label{subsubsec:reach}
Let $\mathrm{reach}(C_r)$ the reach-set of $C_r$ be the set of nodes accessed at least \textit{once} to train \textit{any} model in $C_r$.  We define $\networkquery^{*}_r$ as the representation of $\networkquery_r$ including only nodes in the reach-set. If the underlying representation of $\networkquery_r$ is a graph, the $\networkquery^{*}_r$ representation is an induced subgraph where \textit{both} nodes incident to an edge are in the reach-set. If $\networkquery_r$ is represented as a list, we simply remove elements not in the reach-set.

The encoding size $\cost(\networkquery^*_r)$ measures only the underlying representation accessed for the creation of the $C_r$ task model set. This is a more appropriate measure of the representation cost, where in practice $|\mathrm{reach}(C_r)| << |V|$. For our evaluation, we always report $\mathcal{E}(\networkquery^*_r)$.

\section{Datasets}

We demonstrate our model selection methodology on label prediction tasks of three different online user activity datasets with high-dimensional attributes: beer review history from \ba{}, music listening history from \lfm{}, and movie rating history from \ml{}. 

\small{}
 \begin{table}[t]
 	\centering
 	\resizebox{\boxscale\columnwidth}{!}{
	\begin{tabular}{|c|c|c|c|c|}
		\hline
		Dataset & $|V|$ & $|A|$ & Labels & $|L|$\\\hline
		\lfm{} 20K \cite{mlg2017_13} & 19,990 &  1.2B & 8 & 16628\\
		\ml{} \cite{Harper:2015:MDH:2866565.2827872} & 138,493 & 20M &8&43179 \\ 
		\ba{} \cite{McAuley:2012:LAA:2471881.2472547} & 33,387 & 1.5M &8 & 13079  \\
		\hline
	\end{tabular}}
 	\caption{A summary of datasets in this paper. $|L|$ reports the total number of positive node labels over 8 labelsets.} 
 	\label{tab:data}
 \end{table}
\normalsize{}

\subsection{\lfm{}}
\lfm{} is a social network focused on music listening, logging and recommendation. Previous work collected the entirety of the social network and associated listening history, comparing the social network to alternative network models for music genre label prediction \cite{mlg2017_13}.

Sparse attribute vectors $\vec{a}_i\in A$ correspond to counts of artist plays, where a non-zero element is the number of times user $i$ has played a particular unique artist. \lfm{} also has an explicit `friendship' network declared by users. We treat this as another possible network model, denoted as $E_{\social}$, and evaluate it against others. 

We evaluate the efficiency of network query functions for the label classification task of predicting whether user `$i$' is a listener of a particular genre. A user is a `listener' of an artist if they have at least 5 plays of that artist. A user is a `listener' of a genre if they are a listener of at least 5 artists in the top-1000 most tagged artists with that genre tag, provided by users. We select a subset of 8 of these genre labels (e.g. `dub', `country', `piano'), chosen by guidance of label informativeness from previous work \cite{mlg2017_13}. 
%\footnote{Further details of all label definitions: \cite{brugereicdm2017}}
\subsection{\ml{}} 

\ml{} is a movie review website and recommendation engine. The \ml{} dataset \cite{Harper:2015:MDH:2866565.2827872} contains 20M numeric scores (1-5 stars) over 138K users. 

Sparse non-zero attribute values corresponds to a user's ratings of unique films. We select the most frequent user-generated tag data which corresponds to a variety of mood, genre, or other criteria of user interest (e.g. `inspirational', `anime', `based on a book'). We select 8 tags based on decreasing prevalence (i.e. `horror', `musical', `Disney'), and predict whether a user is a `viewer' of films of this tag (defined similarly to \lfm{} listenership), using attributes and labels sampled by the network query function.

\subsection{\ba{}} 

\ba{} is a website containing text reviews and numerical scores of beers by users. Each beer is associated with a category label (e.g. `American Porter', `Hefeweizen'). We select 8 categories according to decreasing prevalence of the category label. We predict whether the user is a `reviewer' of a certain category of beer (defined similarly to \lfm{} listenership).

\subsection{Network and label sparsity}

When building \knn~and \tth~representations of our data, we construct both `dense' and `sparse' models, according to edge threshold $\rho$. For \lfm{}, we fix $\rho= |E_{\social}|$ for the dense network, and $\rho= 0.5 \times |E_{\social}|$ for the sparse. For both \ba{} and \ml{}, a network density of $0.01$ represents a `dense' network, and $0.0025$ a `sparse'. However, all of these networks are still `sparse' by typical definitions. 

User labels on all three datasets are binary (`this' user is a listener/reviewer of `this' genre/category), and sparse.  We therefore use a label `oracle' and present only positive-label classification problems. This allows us to evaluate only distinguishing listeners etc. rather than learning null-label classifiers where label majority is always a good baseline.

Table \ref{tab:data} ($|L|$ columns) reports the count of non-zero labels over all 8 labelsets. This is the total number of nodes on which evaluation was performed. The \textit{total} number of task models instantiated per network query function is, thus, $b \times |K| \times |L|$. Each local task model can be independently evaluated, allowing us to scale arbitrarily.

\subsection{Validation and testing}

Following \citet{brugereicdm2017}, for each of the three datasets, we split data into temporally contiguous `validation,' `training,' and `testing,' partitions of approximately 1/3 of each dataset. Training is on the middle third, validation is the segment prior, and testing on the latter segment. Model selection is performed on validation, and this model is evaluated on testing. 

%  \begin{table*}[t]
%  	\centering
%  	\resizebox{\bannerwidth\textwidth}{!}{
% 	\begin{tabular}{|c|c|c|c|c|c|}
% 		\hline
% 		Dataset & $|V|$ & $|A|$ & Labelsets & $|L|$ (validation) & $|L|$ (test)\\\hline
% 		\lfm{} 20K \cite{mlg2017_13} & 19,990 &  1,243,483,909 artist plays & 8 &6766& 9862 \\
% 		\ml{} \cite{Harper:2015:MDH:2866565.2827872} & 138,493 & 20,000,263 movie ratings & 8&28364& 14815 \\ 
% 		\ba{} \cite{McAuley:2012:LAA:2471881.2472547} & 33,387 & 1,586,259 beer ratings &8 & 6915& 6164  \\
		
% 		\hline
% 	\end{tabular}}
%  	\caption{A summary of datasets in this paper. $|L|$ reports the total number of positive node labels over 8 labelsets.} 
%  	\label{tab:data}
%  \end{table*}

\section{Evaluation}
In this evaluation, we demonstrate robustness of our sampling strategy for stable model ranking, as well as present model significance testing and sensitivity to noise. 

\subsection{Bootstrap and rank stability}

We evaluate our methodology over $b=20$ samples of varying size $K=[25,50,...150]$ for a total of $b \times |K|$ task instances to evaluate node $i$ on a given label.

\begin{figure}
\centering
  \includegraphics[width=\subfigwidth\columnwidth]{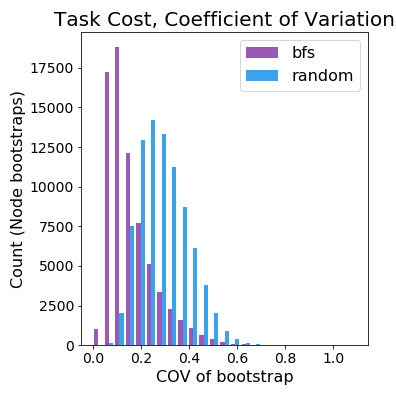}
   \includegraphics[width=\subfigwidth\columnwidth]{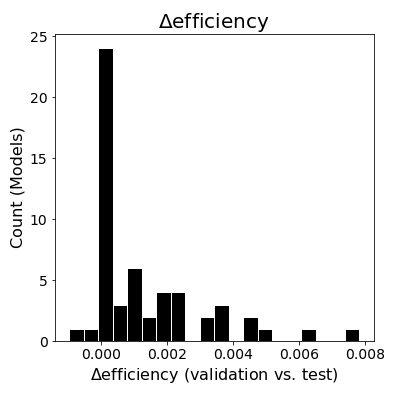}
  \caption{(Left) A distribution of the coefficient of variation ($\mu/\sigma$) over $b=20$ samples for a particular $\classifierset$ set of classification instances, for \lfm{} \social{}. This estimates the median of $b=100$ with low error ($\leq 0.02$). (Right) The change in efficiency between validation and testing partitions: $\mathcal{E}(\networkquery_{r, test})-\mathcal{E}(\networkquery_{r,validation})$.} 
  \label{fig:bootstrap}
\end{figure}

We now test the stability our choice of sampling parameters, over $b$, and across partitions. Figure \ref{fig:bootstrap} (Left) reports the coefficient of variation ($\mu/\sigma$) of task encoding costs for the $b$ classifiers in $\classifierset$. These are smaller for the \social{}-\bfs{} model on \lfm{} than for \social-\random. The median of these distributions at $b=20$ estimates the median at $b=100$ with low error ($\leq 0.02$) for a selection of models verified. Figure \ref{fig:bootstrap} (Right) reports the signed difference in efficiency between models on the test and validation partition. The models with $\Delta$efficiency $> 0.004$ (or an increase of 250 bytes/correct) are all \bfs{} on \ml{}, which may be a legitimate change in efficiency. This shows that our measurements are robust for estimating encoding cost at $k$, and efficiency is stable across partitions.

We test the stability of $\efficiency$ and its correlation to correct predictions. Table \ref{tab:kendall_e_vs_correct} (Top) reports the Kendall's tau rank order statistic measuring correlation between the `efficiency' ranking of models. This further shows stability in the models between validation and testing partitions. Table \ref{tab:kendall_e_vs_correct} (Bottom) in contrast reports the $\tau$ between the ranked models according to efficiency, and according to correct predictions. Within the same partition, this rank correlation is lower than the efficiency rank correlation \textit{across} partitions. This means that efficiency is quite stable in absolute error and relative ranking, and efficiency ranking is not merely a surrogate for ranking by correct prediction. Were this true, we would not need to encode the model. Nor are the ranks of these measures completely \textit{uncorrelated}. 

\begin{table}
\centering
\resizebox{\boxscale\columnwidth}{!}{
\begin{tabular}{|c|c|c|c|}
\hline
\multicolumn{4}{|c|}{$\tau$, $\efficiency$, Validation vs. Test} \\
\hline
&\lfm{} & \ml{} &\ba{} \\
\hline\hline
$\tau$ & 0.89 & 0.61 & 0.82 \\
$p$-value & 1e-8  & 5e-4 & 3e-6 \\
\hline\hline
\multicolumn{4}{|c|}{$\tau$, $\efficiency$ vs. $\taskcorrect$ (validation)} \\
\hline
$\tau$ & 0.73 & 0.38 & 0.70 \\
$p$-value &3e-6  & 0.03 & 7e-5 \\
\hline
\end{tabular}}

\caption{(Top) The Kendall's tau rank correlation coefficient between model efficiency of validation vs. test partitions. (Bottom) $\tau$ rank correlation between efficiency and correct predictions.}
\label{tab:kendall_e_vs_correct}
\end{table}
\subsection{Efficiency features}
Our definition of efficiency yields interpretable features which can characterize models and compare datasets. First, we compare the model and task encoding cost by calculating the ratio of model encoding cost to the total cost of encoding the model and collection of task models. 

Second, recall that in the efficiency definition, we select the most efficient $\kappa_i$ per node $i$, with $\kappa_i \leq 150$ in our evaluation. This gives the model flexibility to search deeper in the query function for each $i$. Nodes with higher $\kappa_i$ are likely harder to classify because the task encoding is typically more expensive with more attribute-vector instances. 

\begin{figure}
\centering
  \includegraphics[width=\figwidth\columnwidth]{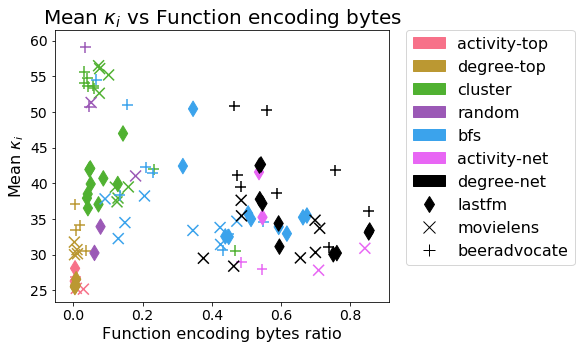}
  \caption{The mean over all $\kappa_i$: $\mathrm{mean}_i(\kappa_i)$ vs. the ratio of model cost by the total encoding cost: $\netcost/(\netcost+\taskcost)$. Colored by network query function, with markers by dataset.} 
  \label{fig:scatter}
\end{figure}

Figure \ref{fig:scatter} compares the ratio of model and total encoding costs (x-axis) vs. the mean over all $\kappa_i$ (y-axis). The models roughly order left-to-right on the x-axis according to Table \ref{tab:functions}. 

All non-network models (\activityflat, \degreeflat, \cluster, \random) are consistently inexpensive vs. their task models, but several also select a higher $\kappa_i$. In contrast, the \degree{} and \activity{} models are consistently costlier because they encode a fixed $m > \max(K)$ in order to sample at each value $k$. All `sorted' models (network and non-network) tend to be most efficient at a lower $\kappa_i$. This may indicate these sorted nodes are more informative per instance, and each additional instance is more costly to incorporate into task models. Finally, \bfs{} has high variance in encoding cost ratio depending on the \textit{reach} of the underlying topology (Section \ref{subsubsec:reach}).

\subsection{Model selection}

We now focus on selecting models from the efficiency ranking in the validation partition. We've already shown that there is high stability in the efficiency ranking between validation and testing partitions, but less correlation between correct predictions and efficiency (Table \ref{tab:kendall_e_vs_correct}). Therefore, in this evaluation, we show we can recover the \textit{best} model in test ($\networkquery_{best}$) with respect to correct predictions, using efficiency selection criteria. 

\begin{table}[ht]
\centering
\resizebox{\boxscalemodelselect\columnwidth}{!}{
\begin{tabular}{|l|c|c|c|}
\hline
\multicolumn{4}{|c|}{Model Selection: $\efficiency$, Evaluation: $\correct(C_r)$} \\
\hline
\multicolumn{1}{|c|}{$\networkquery_{select}$ (validation)}&$\frac{\mathcal{E}(\networkquery_{select})}{\mathcal{E}(\networkquery_{best})}$& $\frac{\cost(\networkquery_{select})}{\cost(\networkquery_{best})}$ & $\frac{\correct(\networkquery_{select})}{\correct(\networkquery_{best})}$ \\
\hline
\hline
\multicolumn{4}{|c|}{\lfm{}, $\networkquery_{best}$: \social-\bfs} \\
\hline

1. \social-\bfs & 1.00 & 1.00 & 1.00 \\
2. \social-\cluster & 0.78 & 0.29 & 0.56 \\
3. \knns-\cluster & 0.72 & 0.08 & 0.50 \\
\hline

\multicolumn{4}{|c|}{\ml{}, $\networkquery_{best}$: \activityflat} \\
\hline

1. \activityflat & 1.00 & 1.00 & 1.00 \\
2. \knns-\bfs &0.12 & 15.79 & 0.28 \\
3. \activity & 0.20 & 130.14 & 0.91 \\
\hline

\multicolumn{4}{|c|}{\ba{}, $\networkquery_{best}$: \activity} \\
\hline

1. \activityflat & 1.20 & 0.01 & 0.86 \\
2. \activity & 1.00 & 1.00 & 1.00 \\
3. \knns-\cluster & 0.22 & 0.09 & 0.20 \\
\hline

\end{tabular}}

\caption{Model selection on `efficiency' ranking in validation ($\networkquery_{select}$, Column 1) compared to the best model in test ranked by correct predictions ($\networkquery_{best}$), for efficiency, encoding cost, and correct prediction ratios (Columns 2,3,4).}
\label{tab:kendall_e}
\end{table}

The left-most column of Table \ref{tab:kendall_e} shows the three top-ranked models ($\networkquery_{select}$) by efficiency in validation. The ratios report the relative performance of the selected model (evaluated in test) for efficiency, total encoding size, and correct predictions.The second column--efficiency ratio--is not monotonically decreasing, because model ranking by efficiency may be different in test than validation (Table \ref{tab:kendall_e_vs_correct}).

The selection by efficiency shows an intuitive trade-off between encoding cost and predictive performance. For \lfm{}, the \social{}-\bfs{} model performs best, but the inexpensive \social{}-\cluster{} model is an alternative. This is consistent with previous results which show \social{}-\bfs{} is much more predictive than network adjacency, and of other underlying network models \cite{mlg2017_13}. 

According to performance ratios, \activity{} is extremely preferred in \ml{} and \ba{}. This measures the extent that the most active users are also most informative in terms of efficiency in these two domains.

\ml{} correctly selects \activityflat{}, and \activity{} is an order of magnitude costlier while maintaining similar correct predictions. On \ba{}, \activity{} performs best on correct predictions, but \activityflat{} is the better model since it preserves $0.86$ of correct predictions but is $0.01$ the encoding cost (it is first ranked by efficiency). This clearly shows why efficiency is preferred to favor parsimonious models.

% \ba{} and \ml{}
% \activity{} is more than an order of magnitude larger than \activityflat expensive model, and the best competing models compensate on encoding cost (e.g. \cluster). 

\begin{figure}
\centering
\includegraphics[width=\threesubfigwidth\columnwidth]{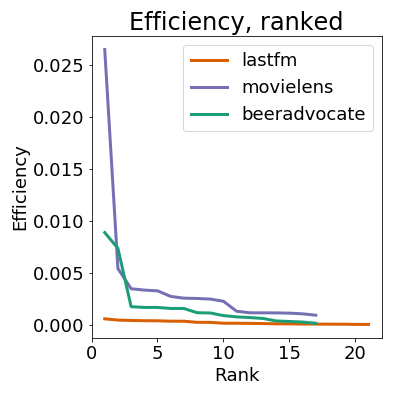}
\includegraphics[width=\threesubfigwidth\columnwidth]{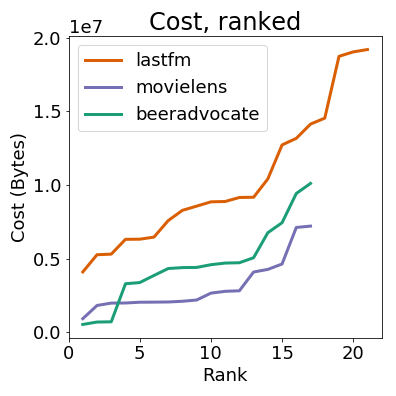}
\includegraphics[width=\threesubfigwidth\columnwidth]{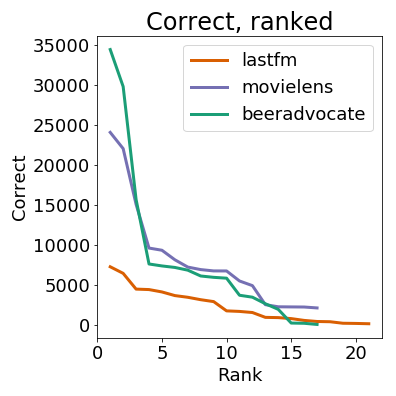}
\caption{Models ranked (x-axis) by efficiency (Left), total encoding cost $\taskcost + \netcost$ (Middle), and the number of correct predictions (Right).}
\label{fig:cost-correct}
\end{figure}

Figure \ref{fig:cost-correct} reports models ranked by efficiency (Left) total encoding cost (Middle), and correct predictions (Right). The \lfm{} ranking has more models because \social{} are included. Ranking by correct predictions shows the extent that the top-ranked \activity{} and \activityflat{} models dominate the correct predictions on \ba{} and \ml{}, which yields low ratios against $\networkquery_{best}$ in Table \ref{tab:kendall_e}. Figure \ref{fig:cost-correct} (Middle) shows that each dataset grows similarly over our set of models, but that each has a different baseline of encoding cost. \lfm{} is particularly costly; the median non-zero attributes per node (e.g. artists listened) is 578, or 7 times larger than \ml{}. These larger attribute vectors yield more expensive task models, requiring more bytes for the same correct predictions. These baselines also yield the same dataset ordering in efficiency. So, the \textit{worst} model on \ml{} (\random{}, 1084 bytes/correct) is more efficient than the \textit{best} model on \lfm{} (\social{}-\bfs{}, 1132 bytes/correct). 

%on a clear preferred model \ml{} and \ba{}{} have a strong model preference   We recover the best model on all three datasets in the top three models. However, there is considerable loss in $\correct()$ for models other than the top. All three datasets recover the best modelWe see significant decrease in

\subsection{Model significance}
In order to compare models with respect to a particular dataset, we measure the \textit{significance} of each model relative to the efficiency over the complete set of models. 

We compare the median difference of efficiency of $\networkqueryr$ to all other models against the median pairwise difference of all models excluding $\networkqueryr$, over the inter-quartile range (IQR) of pairwise difference. This measure is a non-parametric analogue to the $z$-score, where the median $\networkqueryr$ differences deviate from the pairwise expectation by at least a `$\lambda$' factor of IQR. 

Let $e_i = \mathcal{E}(\networkquery_i)$, the efficiency value for an arbitrary network query function $\networkquery_i$, and $\lambda$ a significance level threshold, then for $i = 1...|F|, j = 1...|F|; i, j \neq r$:\small{}
\begin{multline}
\mathrm{significance}(F, r, \lambda) = \\ \frac{\median_i(|e_r - e_i|) - \median_{i,j}(|e_i- e_j|)}{\iqr_{i,j}(|e_i - e_j|)} \geq \lambda 
\label{eq:zscore}
\end{multline}
\normalsize{}This is a \textit{signed} test favoring larger efficiency of $\networkqueryr$ than the expectation. The median estimates of the $e_r$ comparisons and pairwise comparisons are also robust to a small number of other significant models, and like the $z$-score, this test scales with the dispersion of the pairwise differences.

This test only assumes `significant' models are an outlier class in all evaluated models $Q$. We propose a diverse set of possible representations; we use this diversity to treat consistency in the pairwise distribution as a robust null hypothesis: i.e. there is no appropriate model of the data within $Q$. Future work will more deeply focus on measuring this `diversity', and determining the most robust set of models suitable for null modeling. 

At $\lambda=1$ in validation, we find five significant models, corresponding to models reported in Table \ref{tab:kendall_e}: \social{}-\bfs{} on \lfm{} ($=1.30$),  \activityflat{} and \activity{} on \ml{} ($=12.29, 1.21$) and \activityflat{} and \activity{} on \ba{} ($=7.64,~6.00$). 

\subsection{Model stability}

For each of the significant models, we measure the impact of noise on its efficiency and significance. We apply node rewiring on the model we are testing, and leave all other models intact. For each neighbor of a node $i$, we re-wire $i$ to a new node at probability $p$. This is an out-degree preserving randomization, which is appropriate because outgoing edges determine the input for the task model on $i$. Sorting heuristics \activity{} and \degree{} are implemented as ad-hoc networks, where each node $i$ has directed edges to some $m$-sized subset of top-ranked nodes. For each top-ranked node adjacent to $i$, we re-map $i$ to a random node (possibly not in the top-ranked subset) with probability $p$. This again preserves the out-degree of $i$, and reduces the in-degree of top-ranked nodes.

\begin{figure}
\centering
\includegraphics[width=\subfigwidth\columnwidth]{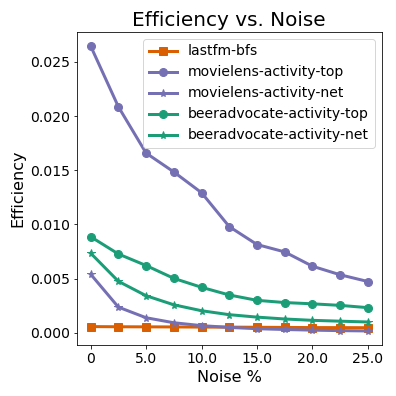}
\includegraphics[width=\subfigwidth\columnwidth]{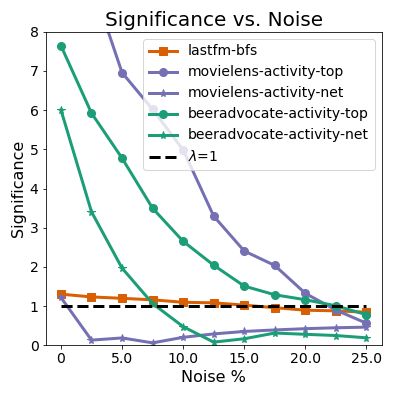}
\caption{Varying level of noise for 5 `significant' models (x-axis), reporting Efficiency (Left) and Significance (Right)}
\label{fig:significance_noise}
\end{figure}

Figure \ref{fig:significance_noise} shows the effect of varying noise $p$ (x-axis), on the efficiency (Left), and significance (Right) of each significant model. \activity{} on both \ml{} and \ba{} quickly lose efficiency under even small noise, and are no longer significant for $\lambda=1$ at $p=0.025$, and $p=0.10$, respectively. \activityflat{} is more robust, remaining significant to $p=0.225$. 

\activity{} is particularly sensitive to noise due to decreased performance in \textit{both} encoding cost and correct predictions. At only $p=0.025$, \activity{} on \ba{} reduces in correct predictions by $15\%$ and \textit{increases} in encoding cost by $31\%$. The encoding cost greatly increases because the cardinality of the set of unique numbers in the \activity{} representation is small (bound by the size of the top-ranked sample). When random nodes are added in the representation, they are near-unique values, greatly increasing the set cardinality and reducing the compression ratio. \activityflat{} shows a similar increase, but proportional to nodes rather than edges.

Finally, \social{}-\bfs{} is easily the most robust to noise. From a lower baseline, it remains significant to $p=0.15$. It loses only $35\%$ of its significance value at any noise level, while all other methods lose $>90\%$. This demonstrates that network models have robustness which might be desirable for further criteria in model selection. For example, our full methodology can be used to select on efficiency significance at some noise level $p$.

\section{Conclusions and Future Work}

In this paper, we have presented a minimum description length approach to perform model selection of networks. We have generalized networks as one particular representation of network query functions used to solve predictive tasks. This formulation allows comparing against arbitrary underlying representations and testing whether a network model is necessary in the first place. 
We propose a general efficiency measure for evaluating model selection for particular tasks on networks, and we show stability for node sampling and model ranking, as well as significance testing and sensitivity analysis for selected models. In total, this methodology is general and flexible enough to evaluate most networks inferred from attributed/labeled data, as well as networks given explicitly by the application of interest against alternative models.

There are several avenues for improving this work. First, exploiting model similarity and prioritizing novel models may more quickly  discover significant models and estimate their value against the full null model enumeration. Furthermore, we aim to study \textit{which} and \textit{how many} models yield a robust baseline null distribution of efficiency measurements for model selection.

There are also extensions to this work. Currently, we learn node task models for each label instance. However, how many task models are needed on our network and can we re-use local models? Efficiency naturally measures the trade-off between eliminating task models vs. the loss in correct predictions. The aim is then task model-set  \textit{summarization} for maximum efficiency. 

%econd, mixed strategies could allow model selection at the \textit{node} level, which further adapts to the `hardness' of a particular node's task. These selected node models could serve as features for \textit{role discovery} or node characterization.  

%full model enumeration. more quickly discovering  comparing model similaritypruning our significance testing procedure could guide the model ordering for evaluation, to more quickly discover significant models according to the characteristics/parameters of non-significant models.

%to order nodes We've generalized tasks on networks to sampling on a set of network query functions with varying underlying implementations and model encoding costs. 

% \section{Problem Specification.}In this paper, we consider the solution of the $N \times
% N$ linear
% system
% \begin{equation} \label{e1.1}
% A x = b
% \end{equation}
% where $A$ is large, sparse, symmetric, and positive definite.  We consider
% the direct solution of (\ref{e1.1}) by means of general sparse Gaussian
% elimination.  In such a procedure, we find a permutation matrix $P$, and
% compute the decomposition
% \[
% P A P^{t} = L D L^{t}
% \]
% where $L$ is unit lower triangular and $D$ is diagonal.
\bibliographystyle{ACM-Reference-Format}
\small{}
\bibliography{acmsmall-sample-bibfile}

\end{document}